\documentclass[letterpaper, 10 pt, conference]{ieeeconf}  % Comment this line out if you need a4paper

\IEEEoverridecommandlockouts                              % This command is only needed if 
\usepackage{amsmath,amsfonts}
\usepackage{algorithm}
\usepackage{algpseudocode}
\usepackage{array}
\usepackage[caption=false,font=normalsize,labelfont=sf,textfont=sf]{subfig}
\usepackage{textcomp}
\usepackage{stfloats}
\usepackage{url}
\usepackage{verbatim}
\usepackage{graphicx}
\usepackage{cite}
\usepackage{xcolor}
\usepackage{url}
\usepackage{cite}
\usepackage{balance}
\usepackage{acronym}
\acrodef{MPPI}{Model Predictive Path Integral}
\acrodef{GNC}{Graduated Non-Convexity}
\acrodef{LSE}{log-sum-exp}
\acrodef{SDP}{Semidefinite Program}
\acrodef{KernelSOS}{Kernel Sum of Squares}
\acrodef{CMA-ES}{Covariance Matrix Adaptation Evolution Strategy}
\acrodef{CEM}{Cross-Entropy Method}
\acrodef{BO}{Bayesian optimization}
\acrodef{RKHS}{reproducing kernel Hilbert space}
\acrodef{NLL}{negative log-marginal likelihood}

\title{\LARGE \bf
Global Sampling-Based Trajectory Optimization for \\ Contact-Rich Manipulation via KernelSOS
}

\author{Zhongqi Wei$^{1}$ and Frederike Dümbgen$^{1,2}$
\thanks{$^{1}$Department of Mechanical Engineering, Carnegie Mellon University,
Pittsburgh, PA 15213, USA {\tt\small \{zhongqi2, fdumbgen\}@andrew.cmu.edu}. $^2$Part of this work was performed while Frederike Dümbgen was with Inria, Département d'informatique de l'ENS, CNRS, PSL Research University, Paris, France. This work was supported by the European Union, through the Horizon Europe research and innovation program under the Marie Skłodowska-Curie (GA no.101207106), and by the Department of Mechanical Engineering, Carnegie Mellon University.}
}
\begin{document}
%\begin{acronym}

%\end{acronym}
\maketitle
\thispagestyle{empty}
\pagestyle{empty}

%%%%%%%%%%%%%%%%%%%%%%%%%%%%%%%%%%%%%%%%%%%%%%%%%%%%%%%%%%%%%%%%%%%%%%%%%%%%%%%%
\begin{abstract}
Contact-rich manipulation is challenging due to its high dimensionality, the requirement for long time horizons, and the presence of hybrid contact dynamics. Sampling-based methods have become a popular approach for this class of problems, but without explicit mechanisms for global exploration, they are susceptible to converging to poor local minima. %, as they perform local stochastic optimization without explicit mechanisms for global exploration. 
In this paper, we introduce Global-MPPI, a unified trajectory optimization framework that integrates global exploration and local refinement. At the global level, we leverage kernel sum-of-squares optimization to identify globally promising regions of the solution space. To enable reliable performance for the non-smooth landscapes inherent to contact-rich manipulation, we introduce a graduated non-convexity strategy based on log-sum-exp smoothing, which transitions the optimization landscape from a smoothed surrogate to the original non-smooth objective. Finally, we employ the model-predictive path integral method to locally refine the solution. We evaluate Global-MPPI on high-dimensional, long-horizon contact-rich tasks, including the PushT task and dexterous in-hand manipulation. Experimental results demonstrate that our approach robustly uncovers high-quality solutions, achieving faster convergence and lower final costs compared to existing baseline methods.
\end{abstract}

% enabling both global coverage and local convergence.
% %%%%%%%%%%%%%%%%%%%%%%%%%%%%%%%%%%%%%%%%%%%%%%%%%%%%%%%%%%%%%%%%%%%%%%%%%%%%%%%%
% A conclusion section is not required. Although a conclusion may review the main points of the paper, do not replicate the abstract as the conclusion. A conclusion might elaborate on the importance of the work or suggest applications and extensions. 
\section{Introduction}

Contact-rich manipulation, such as pushing, grasping, and in-hand manipulation, is a fundamental task in modern robotics~\cite{elguea2023review}. These tasks typically involve high-dimensional, underactuated systems and hybrid contact dynamics, resulting in highly non-convex, non-smooth optimization landscapes. As a result, computing long-horizon optimal plans and control policies for contact-rich tasks remains a core challenge.

To address this challenge, sampling-based optimization has recently received significant attention~\cite{xue2025full}\cite{rozzi2024combining}. By relying solely on function evaluations, these methods can handle highly nonlinear, non-smooth contact dynamics that induce complex cost landscapes, without requiring higher-order information such as gradients. In addition, enabled by recent advances in GPU-accelerated parallel simulation, sampling-based methods can efficiently evaluate large numbers of candidate control trajectories via massive parallel rollouts~\cite{alvarez2025real}. These aspects make sampling-based methods particularly appealing for the contact-rich trajectory optimization problem.

\begin{figure}[!th]
    \centering
    \includegraphics[width=1\linewidth]{figs/overview.png}
    \caption{Overview of Global-MPPI. Our approach consists of three coupled components. First, we introduce a graduated non-convexity strategy via log-sum-exp (LSE) smoothing to transition the optimization landscape from a smoothed surrogate to the original non-smooth objective, enabling coarse-to-fine global exploration. Second, KernelSOS leverages samples to construct a global non-parametric surrogate and proposes a global solution candidate via semidefinite programming. Finally, the model-predictive path integral (MPPI) method performs efficient local refinement. Together, these components operate in a coarse-to-fine manner to balance
global exploration and local convergence in contact-rich manipulation problems.}
    \label{overview}
\end{figure}
Popular sampling-based approaches include \ac{MPPI}~\cite{gandhi2021robust}, \ac{CEM}~\cite{pinneri2021sample}, and \ac{CMA-ES}~\cite{hansen_cma_2023}. These methods perform iterative stochastic optimization by sampling candidate trajectories and updating the solution based on the observed performance of those samples. A recent approach, DIAL-MPC~\cite{xue2025full}, extends \ac{MPPI} by incorporating annealing strategies to balance coverage and convergence, demonstrating promising performance on contact-rich problems. However, all of these methods are primarily local in nature~\cite{jordana2025introduction}, often leading to suboptimal convergence or task failure in contact-rich settings. This limitation becomes pronounced as task complexity, planning horizon, and system dimensionality increase. To improve global exploration, regularization terms that maximize entropy during sampling can be added to such formulations, yielding better exploration properties~\cite{aoyama_generalized_2024,haarnoja_reinforcement_2017}.%,ziebart_modeling_2010}.

This paper highlights the potential of a recent global optimization method, \ac{KernelSOS}~\cite{rudi2020finding}, to overcome those limitations. KernelSOS provides a means to identify globally promising regions in non-parametric optimization settings. Unlike local sampling-based methods, KernelSOS constructs a global surrogate of the cost landscape from samples and performs global minimization of this surrogate via solving a \ac{SDP}. KernelSOS has been previously studied in domains such as optimal control and estimation in robotics~\cite{berthier_infinite_2022,groudiev2025sampling}, but its application to high-dimensional, long-horizon contact-rich problems has been out of reach. 

In particular,~\cite{groudiev2025sampling} proposes a model-based first-order local solver to refine the coarse solution candidate, but fast and reliable local solvers are harder to come by in contact-rich problems. The method also requires extensive manual tuning for the problem class. However, for long-horizon tasks such as contact-rich manipulation, the cost landscape can vary significantly due to frequent contact mode switches and discontinuous dynamics, and there may not be a single optimal set of hyperparameters. Finally, the high non-convexity of contact-rich tasks, the sparse information content (large areas with no gradient where no contact happens), and difficult-to-discover local minima are a challenge for global sampling methods like \ac{KernelSOS}. As a result, identifying promising regions of the search space may require prohibitively large numbers of samples, leading to computationally intractable \ac{SDP} problems.
% which may require prohibitively large numbers of samples, leading to prohibitively costly \ac{SDP}s, to even detect regions of interest. 

% we could talk more concretely about the shortcomings of the above approach:
% - uses a model-based method as a local method; hard for difficult-to-model tasks like contact-rich manipulation
% - manual tuning *per* task, makes it harder to use

We introduce Global-MPPI, a global sampling-based trajectory optimization framework for contact-rich problems, which addresses these challenges one by one. %. We integrate \ac{KernelSOS} for global exploration and \ac{MPPI} for local refinement within a unified optimization pipeline. Since \ac{KernelSOS} provides strong global guarantees for the smooth cost functions~\cite{rudi2020finding}, we also incorporate a strategy based on \ac{GNC}, which gradually increases non-convexity to enable coarse-to-fine exploration of the optimization landscape. We evaluate our method on contact-rich benchmark tasks and show that it achieves faster convergence and lower costs than state-of-the-art sampling-based baselines. 
Our main contributions are as follows. 

\begin{enumerate}
    \item \textbf{A fully sampling-based framework}: Global-MPPI integrates KernelSOS for global exploration and \ac{MPPI} for local refinement. Both are sampling-based and thus do not require differentiable contact modeling, making them suitable for integrating complex and accurate physics simulators. %within a unified pipeline tailored to contact-rich trajectory optimization problems.
    \item \textbf{Graduating smoothing}: We introduce a \ac{GNC} strategy based on \ac{LSE} smoothing, which provides a general mechanism for iteratively smoothing the optimization landscape, improving the exploration capabilities of KernelSOS.
    \item \textbf{Auto calibration for KernelSOS}: We propose a calibration method that automatically and continuously adapts appropriate kernel parameters in KernelSOS based on maximizing the marginal likelihood of the function fit.
    \item \textbf{Validation on contact-rich tasks}: We validate the performance of the proposed method on contact-rich manipulation tasks based on a GPU-accelerated physics simulator, including the PushT task and dexterous in-hand manipulation. The open-source implementation is made publicly available.\footnote{Code will be released upon acceptance of the manuscript.}
\end{enumerate}

This paper is organized as follows. After reviewing contact-rich problems and sampling-based optimization methods for robotics in Sec.~\ref{Related Work}, we formally state the problem formulation for contact-rich trajectory optimization in Sec.~\ref{Problem Statements}. We then detail the proposed method in Sec.~\ref{sec:method} and present numerical experiments in Sec.~\ref{experiments}. We conclude with a discussion of the results and limitations in Sec.~\ref{Conclusion}.

\section{Related Work}
\label{Related Work}
\subsection{Contact-rich Optimization in Robotics}

In recent years, learning-based methods, such as deep reinforcement learning and behavior cloning, have demonstrated impressive robustness in contact-rich tasks such as real-world quadruped locomotion~\cite{schneider2024learning} and robotic manipulation~\cite{han2023survey}~\cite{yu2022dexterous}. However, these methods typically require careful hyperparameter tuning and initialization, time-consuming training, and, in some cases, extensive data collection. The resulting policies are heavily dependent on the training setup, limiting their generalization to unseen tasks and environments.

%\subsection{Model-based methods}
The above methods implicitly handle complex contact interactions using stochastic approximations of system dynamics~\cite{suh2022differentiable}. An alternative line of work exploits these contact models and approximations more explicitly in optimal control frameworks~\cite{suh2022bundled, pang2023global,le_lidec_leveraging_2024}, sometimes using complementarity-based formulations~\cite{Posa2014, carpentier_compliant_2024, kurtz2026inverse}. Although they have demonstrated promising performance, these methods are inherently local and often require high-quality initial guesses, which are difficult to obtain in practice. Alternatively, contact-rich problems have also been studied through global optimization formulations. These methods typically cast the problem as a combinatorial optimization problem, which can be solved using mixed-integer programming or via convex relaxation~\cite{8141917},~\cite{graesdal2024towards}. Although such approaches can provide optimality guarantees, they do not scale with dimension and the number of contact modes and have therefore been limited to relatively simple systems.

\subsection{Sampling-based Methods}
Sampling-based or zero-order methods have been widely applied in robotics due to their flexibility in handling arbitrary non-smooth dynamics and constraints, as well as parallelizability~\cite{xue2025full}. These approaches rely solely on function evaluations and do not require gradient or higher-order derivative information, making them well-suited for incorporating non-differentiable simulators or real-world rollouts. Prominent methods such as \ac{MPPI} and \ac{CMA-ES} perform stochastic optimization by repeatedly sampling around a nominal solution and updating the sampling distribution~\cite{williams2017model}. Although robust to non-smooth objectives, those methods remain fundamentally local in nature. \ac{BO}~\cite{berkenkamp2023bayesian,shahriari_taking_2016,jones_efficient_1998}, on the other hand, incorporates global exploration by fitting a surrogate model of the objective and selecting new samples by maximizing an acquisition function that trades off exploration and exploitation. While the acquisition function is typically easier to optimize than the original objective, its maximization remains a non-convex problem and may be prone to local maxima. A recent contender among zero-order methods is KernelSOS~\cite{rudi2020finding}. Like \ac{BO}, KernelSOS constructs a global surrogate model within a prescribed function class. However, instead of iteratively optimizing an acquisition function, KernelSOS directly minimizes the surrogate in a single step via solving an \ac{SDP}, providing global optimality guarantees within the surrogate space. %This one-shot formulation avoids repeated non-convex acquisition function optimization and is particularly well-suited for objectives that are expensive to evaluate or lack smoothness.

\subsection{Graduated Non-Convexity}
\acf{GNC} is a widely used strategy for optimizing non-convex objective functions and has been successfully applied across domains, such as computer vision~\cite{yang2020graduated} and machine learning~\cite{rose2002deterministic}. Existing \ac{GNC} formulations, however, are often tailored to specific loss functions. For example, Yang et al.~\cite{yang2020graduated} propose dedicated \ac{GNC} strategies for the Geman–McClure and truncated least-squares losses. While effective, such formulations are inherently function-specific and therefore lack generality for arbitrary non-convex cost functions.

\vspace{1em}
Motivated by these limitations, we propose a global, training-free, and non-parametric trajectory optimization framework for contact-rich problems. Our approach integrates a general, function-agnostic \ac{GNC} strategy with KernelSOS for global optimization and MPPI for local refinement, both of which are sampling-based and particularly well-suited to contact-rich settings with non-convex, non-smooth cost landscapes. %Unlike prior methods that rely on gradient-based local solvers~\cite{groudiev2025kernelsos}, our local refinement is performed entirely via sampling, making it particularly well-suited to contact-rich settings with non-convex, non-smooth cost landscapes.
\section{Problem Statement}
\label{Problem Statements}

The problem of interest in this paper is a finite-horizon trajectory optimization problem using a contact-implicit representation, where contact forces are implicitly encoded in the system dynamics:
\begin{equation}\label{eq:gcs}
\begin{aligned}
\min_{\mathbf{u}} \quad 
& \mathcal{J}(\mathbf{u}) := \sum_{t=0}^{T-1} \ell_t(x_t, u_t)
+ \ell_T(x_T) \\
\text{s.t.} \quad
& x_0 = x_{\mathrm{init}}, \\
& x_{t+1} = f_{\mathrm{dyn}}(x_t, u_t), \quad t = 0, \dots, T-1 , \\
%& x_{t} \in \mathcal{X}, \quad t = 0, \dots, T , \\
& u_{t} \in \mathcal{U}, \quad t = 0, \dots, T-1.
\end{aligned}
\end{equation}

Here, $x_t$ and $u_t$ denote the state and control input at time step $t$ and $T$ is the planning horizon. The control sequence is denoted by $\mathbf{u}= \{u_0, u_1, \dots, u_{T-1}\}$. The functions $\ell_t(\cdot)$ and $\ell_T(\cdot)$ represent the stage cost and the terminal cost, respectively, $f_{\mathrm{dyn}}(\cdot)$ denotes the contact-implicit dynamics, and $\mathcal{U}$ is the feasible control set. In practice, we employ the single-shooting formulation, in which the unknown state trajectory and dynamics constraints are eliminated by rolling out the system dynamics $f_{\mathrm{dyn}}(\cdot)$ starting from the initial condition $x_\mathrm{init}$. The problem thus simplifies to computing a feasible control sequence $\mathbf{u}$ that minimizes the resulting cumulative cost~$\mathcal{J}(\mathbf{u})$.

This problem is challenging due to several factors: First, the contact-implicit dynamics $f_{\mathrm{dyn}}(\cdot)$ introduces non-smooth and highly non-convex behavior as contact modes change. As a result, the cost landscape often exhibits sharp, asymmetric basins and pronounced non-smoothness, which hinder efficient global exploration and may cause local methods to get trapped in suboptimal solutions. Second, as the task dimensionality increases, the search space of the control sequence $\mathbf{u}$ grows exponentially, substantially increasing the computational burden and the difficulty of achieving global coverage. Finally, successful manipulation typically requires long horizons, thereby further increasing the problem's dimensionality.
\section{Method}
\label{sec:method}

To address the challenges of contact-rich trajectory optimization discussed in Section~\ref{Problem Statements}, we introduce Global-MPPI, a global sampling-based trajectory optimization framework tailored for contact-rich problems. We start by providing an overview of the algorithm and methodological insights to explain why Global-MPPI enables effective global optimization over non-smooth landscapes. We then present technical details for each component of the algorithm, along with practical considerations for its successful application to contact-rich tasks.

\begin{figure*}[!t]
    \centering
    \includegraphics[width=\textwidth]{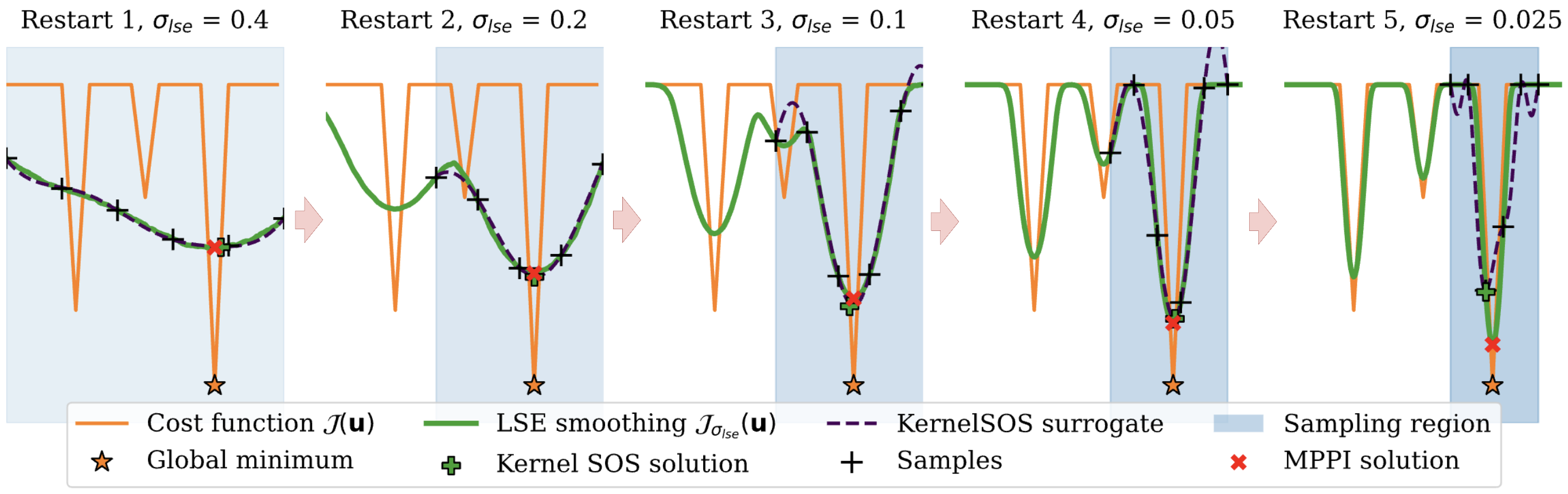}
    \caption{Illustration of the advantages of Global-MPPI. For contact-rich problems, the original cost function (orange curve) often has sharp, asymmetric peaks. Graduated non-convexity via log-sum-exp smoothing transitions the optimization landscape from a smoothed surrogate (green curve) to the original non-smooth objective at each restart stage. This graduating smoothing facilitates global approximation by KernelSOS and enlarges the basin of attraction of low-cost solutions. MPPI then locally refines the solution from KernelSOS. As a result, the final solution gradually converges to the true global minimum.}
    \label{fig:lse}
\end{figure*}

\subsection{Overview of Global-MPPI}

Global-MPPI is a global sampling-based trajectory optimization framework designed to address contact-rich tasks. Fig. \ref{fig:lse} provides a univariate example to illustrate its high-level idea. Our approach integrates three coupled components that enable reliable global exploration and local convergence toward the global optimum (orange star) in highly non-smooth, non-convex cost functions (orange curve). First, we employ a \ac{GNC} strategy based on \ac{LSE} smoothing. By gradually decreasing the smoothing parameter $\sigma_{lse}$, the optimization landscape transitions from a smoothed convex function to the original non-smooth cost function (green curve), allowing coarse-to-fine global exploration. Second, at each stage, we leverage KernelSOS to identify the global solution (green cross) through surrogate minimization (dashed curve) via solving an \ac{SDP}. Crucially, KernelSOS operates on the smoothed cost function, which, as we will show, leads to better performance than if we were to use the original function evaluations. Finally, we employ MPPI as a local optimizer, initialized with the global solution candidate from KernelSOS, to perform local refinement (red cross) of the LSE-smoothed surrogate function. Together, these components enable Global-MPPI to systematically explore highly non-smooth, non-convex cost landscapes, avoid local minima, and progressively converge to high-quality solutions in non-parametric, contact-rich settings.

\subsection{Graduated Non-Convexity via Log-Sum-Exp Smoothing}

To address non-convex and non-smooth optimization landscapes, we adopt a \ac{GNC} approach that solves a sequence of surrogate optimization problems with gradually increasing non-convexity, enabling coarse-to-fine exploration of the optimization landscape. The key idea of GNC is to first optimize a strongly smoothed, convex surrogate of the original problem, which is easier to optimize, and then gradually increase the level of non-convexity until the original non-convex objective is recovered. The solution obtained at each stage is used to warm-start the subsequent optimization stage. This continuation process effectively enlarges the basin of attraction of low-cost solutions and improves robustness with respect to initialization and sampling variability. Moreover, the theoretical guarantees of KernelSOS rely on smoothness assumptions on the objective function~\cite{rudi2020finding}. The smoothed functions produced by GNC make it easier for KernelSOS to fit accurate surrogates and identify high-quality solutions compared to the original non-smooth objective.

In this work, we propose a general \ac{GNC} strategy based on \ac{LSE} smoothing~\cite{jordana2025introduction}, which is applicable to arbitrary non-convex, non-smooth functions. Given a cost function $\mathcal{J}(\mathbf{u})$, the LSE-smoothed surrogate function $\mathcal{J}_{{lse}}(\mathbf{u}, \sigma_{lse})$ is defined as
\begin{equation}
\label{eq:lse_obj}
\mathcal{J}_{{lse}}(\mathbf{u}, \sigma_{lse})
=
-\lambda \log \left(
\mathbb{E}\left[
\exp\left(
-\frac{1}{\lambda} \mathcal{J}(\mathbf{u} + \sigma_{lse} \boldsymbol{\epsilon})
\right)
\right]
\right),
\end{equation}
where $\boldsymbol{\epsilon}$ contains \emph{i.i.d.} samples $\epsilon_i \sim \mathcal{N}(0, I)$, $\lambda > 0$ is temperature parameter and $\sigma_{lse} > 0$ is a GNC smoothing parameter. 

As illustrated in Fig.~\ref{fig:lse}, large values of $\sigma_{lse}$ produce a strongly smoothed, nearly convex surrogate, while small values recover the original non-smooth cost function. By gradually reducing $\sigma_{lse}$ across restart phases, the optimization landscape transitions from a smoothed, nearly convex surrogate to the original cost function. Moreover, \ac{LSE} smoothing places greater weight on global minima and attenuates local minima, thereby enlarging the basins of attraction around high-quality solutions. This effect facilitates an effective global exploration of the solution space in our framework.

\subsection{KernelSOS for Global Optimization}

Importantly, we expect the smoothed function to still exhibit non-convexity, and local solvers would be at risk of getting trapped in local minima. Instead, we use \ac{KernelSOS}, which is a global sampling-based optimization method introduced in \cite{rudi2020finding}. The key idea of KernelSOS is to generalize classical sum-of-squares global optimization by replacing fixed polynomial bases with kernel functions, enabling a convex relaxation that allows to retrieve approximate global minima of general, non-parametric objectives. Since it only uses finite function and kernel evaluations, it is suitable for sampling-based optimization pipelines for contact-rich problems.

We define the kernel matrix $K_\sigma = \bigl(k_\sigma(\mathbf{u}^{(i)}, \mathbf{u}^{(j)})\bigr)_{i,j \in [N]}$, where $k_\sigma$ is a chosen positive-definite kernel, $\sigma$ is the kernel parameter and $N$ is the finite number of samples. We denote its Cholesky decomposition by $K_{\sigma} = R^\top R$, with $R$ an upper-triangular matrix. KernelSOS constructs the following \ac{SDP}:
\begin{equation}
\begin{aligned}
c^{*}_{ksos}
\;=\;
\max_{c \in \mathbb{R},\, B \succeq 0}
&\quad c - \mu \operatorname{Tr}(B) \\
\text{s.t.}
&\quad \mathcal{J}(\mathbf{u}^{(i)}) - c = r_i^\top B r_i,
\quad i \in [N],
\end{aligned}
\label{eq:kernel_sos}
\end{equation}
where $r_i$ denotes the $i$-th column of $R$, $\operatorname{Tr}$ is the trace operator, $\mu$ is a regularization parameter, and $B \succeq 0$ is a positive semidefinite matrix that parameterizes the kernelized sum-of-squares surrogate. The resulting KernelSOS surrogate function is defined as $f(\mathbf{u}) = c + v(\mathbf{u})^{\top} R^{-1} B R^{-\top} v(\mathbf{u})$, where $v(\mathbf{u}) = \bigl(k_\sigma(\mathbf{u}^{(1)}, \mathbf{u}), \ldots, k_\sigma(\mathbf{u}^{(N)}, \mathbf{u})\bigr)$ is the kernel feature vector constructed from the samples $\{\mathbf{u}^{(i)}\}_{i=1}^N$.

As shown in~\cite{rudi2020finding}, the relaxation gap between the solution of ~\eqref{eq:kernel_sos} and the true global optimum $c^\star$ satisfies $c_{ksos}^\star \leq c^\star$ if $\mathcal{J}(\mathbf{u}) - c$ lies in the chosen \ac{RKHS}. In this case, the relaxation gap is guaranteed to asymptotically approach zero as $N \to \infty$. 

In practice, the magnitude of the relaxation gap thus depends on the choice of kernel and its associated hyperparameters, since they define the \ac{RKHS}. Without careful calibration of the kernel parameters, the KernelSOS surrogate function may fail to accurately approximate the original objective $\mathcal{J}(\mathbf{u})$. Furthermore, there may not be a single function class that accurately approximates the landscape over the long-horizon task, particularly as we switch between contact modes. Therefore, we propose to automatically select the kernel parameter in KernelSOS, using a data-driven linear-to-quadratic auto-calibration strategy based on an empirical Bayes approach. 

Specifically, we estimate the kernel parameter by minimizing the \ac{NLL} with respect to the linear Gaussian process surrogate model. This marginal likelihood maximization provides a statistically principled, computationally efficient, and numerically well-conditioned approach to kernel parameter estimation~\cite{murphy2012machine}. Although KernelSOS uses a quadratic surrogate, we found that the optimal kernel identified using the linear surrogate, which is significantly easier to obtain, transferred well enough for our purposes. %Since the linear surrogate from the Gaussian process and the quadratic sum-of-squares surrogate from KernelSOS are both constructed within the same RKHS, the calibrated kernel parameter can be directly transferred to the sum-of-squares surrogate without additional tuning. 

Given a set of sampled inputs $U = [\mathbf{u}^{(1)}, \mathbf{u}^{(2)}, \ldots, \mathbf{u}^{(N)}]^\top$ and their corresponding cost evaluations $y = [\mathcal{J}_{lse}(\mathbf{u}^{(1)}), \ldots, \mathcal{J}_{lse}(\mathbf{u}^{(N)})]^\top$, we estimate the kernel parameters by minimizing the NLL:
\begin{equation}
\begin{aligned}
\mathrm{NLL}(\sigma)
& = 
-\log p(y \mid U) \\
& = 
\frac{1}{2} y^\top (K_\sigma)^{-1} y
+
\frac{1}{2} \log \lvert K_\sigma \rvert
+
\frac{N}{2} \log (2\pi)
\end{aligned}
\end{equation}

Since $\mathrm{NLL}(\sigma)$ is generally non-convex, we adopt a two-stage kernel selection strategy that first performs a coarse grid search over a predefined candidate set $\mathcal{S}$, followed by local gradient-based refinement to minimize the NLL. The calibrated kernel parameter is $\sigma_{ksos} \;=\; \arg\min_{\sigma \in \mathcal{S}} \, \mathrm{NLL}(\sigma)$. This automatic linear-to-quadratic calibration strategy leverages linear models to select kernel parameters, eliminating the need for ad hoc heuristics and exhaustive hyperparameter tuning.

% The key intuition is that the linear model provides a reliable estimate of the appropriate kernel bandwidth from data, while the quadratic surrogate in KernelSOS captures richer curvature information required for global optimization. By decoupling kernel calibration from higher-order surrogate fitting, this strategy improves robustness and avoids overfitting that may arise when directly tuning the quadratic model.

% \textcolor{red}{add more explicitly linear to = calibration}

% The NLL provides a principled criterion for kernel calibration by quantifying how well the kernel-induced RKHS explains finite function evaluations. Although the NLL is typically derived under a Gaussian process formulation, it can be interpreted as a kernel-dependent evaluation functional for linear function classes in the RKHS~\cite{murphy2012machine}. Since KernelSOS constructs surrogates directly in this RKHS, the kernel parameter determines the expressiveness of the function space in which the sum-of-squares constraints are enforced. Minimizing the NLL therefore provides an automatic, data-driven mechanism for kernel parameter calibration.

\subsection{MPPI for local refinement}
While KernelSOS can discover globally promising solutions across the entire action space, it is less accurate than local solvers in high-dimensional settings. In particular, due to the curse of dimensionality and the limited number of samples, the KernelSOS surrogate may fail to accurately approximate the original cost function, leading to suboptimal solutions. In contrast, local solvers are computationally efficient and well-suited to refine solutions within a basin of attraction once a high-quality initialization is
available.

In our framework, we employ \ac{MPPI} as a local refinement method. MPPI is particularly well-suited for this role because its sampling-based update can be interpreted as a natural stochastic gradient-descent step on the LSE-smoothed objective~\cite{jordana2025introduction}, which naturally aligns with the \ac{GNC} scheme adopted in our method. In particular, the gradient of the smoothed objective in~\eqref{eq:lse_obj} satisfies:
\begin{equation}
\label{eq:lse_grad}
\nabla \mathcal{J}_{{lse}}(\mathbf{u}, \sigma_{lse})
=
-\frac{\lambda}{\sigma_{lse}}
\frac{
\mathbb{E}\left[
{\boldsymbol{\epsilon}}
\exp\!\left(
-\frac{1}{\lambda}
\mathcal{J}(\mathbf{u}+\sigma_{lse}{\boldsymbol{\epsilon}})
\right)
\right]
}{
\mathbb{E}
\left[
\exp\!\left(
-\frac{1}{\lambda}
\mathcal{J}(\mathbf{u}+\sigma_{lse}{\boldsymbol{\epsilon}})
\right)
\right]
}
\end{equation}
Therefore, a stochastic gradient update yields:
\begin{equation}
\label{eq:mppi_gd_general}
\mathbf{u}_{k+1}
=
\mathbf{u}_k
-\alpha \nabla \mathcal{J}_{{lse}}(\mathbf{u}_k, \sigma_{lse})
\approx
\mathbf{u}_k
+
\alpha \frac{\lambda}{\sigma_{lse}}
\sum_{i=1}^{N} w_i\,\boldsymbol{\epsilon}^{(i)} .
\end{equation}
Here, $\alpha$ denotes the step size, 
the normalized weights $w_i$ are defined as:
\begin{equation}
w_i
=
\frac{
\exp\!\left(
-\frac{1}{\lambda}\, \mathcal{J}\!\left(\mathbf{u}_k + \sigma_{lse}\boldsymbol{\epsilon}^{(i)}\right)
\right)
}{
\sum_{j=1}^{N}
\exp\!\left(
-\frac{1}{\lambda}\, \mathcal{J}\!\left(\mathbf{u}_k + \sigma_{lse}\boldsymbol{\epsilon}^{(j)}\right)
\right)
},
\end{equation}
Choosing $\alpha = \sigma_{lse}^{2}/\lambda$ and noting that $\sum_{i=1}^{N} w_i=1$, we recover the standard MPPI update rule:
\begin{equation}
\label{eq:mppi_update}
\mathbf{u}_{k+1}
\approx
\mathbf{u}_k
+
\sigma_{lse}
\sum_{i=1}^{N} w_i\,{\boldsymbol{\epsilon}^{(i)}}
=
\sum_{i=1}^{N} w_i
\left(\mathbf{u}_k + \sigma_{lse}{\boldsymbol{\epsilon}^{(i)}}\right).
\end{equation}

Overall, because MPPI relies solely on cost evaluations and does not require analytic models or gradients, it serves as an effective local refinement method in contact-rich environments.

\subsection{Final Algorithm and Implementation Details}
Having introduced the key algorithmic components, we formally present our methods in Algorithm~\ref{alg:kernelSOS-MPPI}. The algorithm operates in a receding-horizon setting and alternates between global exploration and local refinement across a sequence of \ac{GNC} restart stages. At each stage, the sampling radius and LSE smoothing parameter $\sigma_{lse}$ are gradually reduced to enable a coarse-to-fine global exploration and local convergence. To enforce trajectory smoothness and reduce the dimensionality of the optimization problem, we adopt a spline-based control parameterization mentioned in~\cite{alvarez2025real}, in which trajectories are represented by cubic splines over a set of control points. The input constraints are incorporated by clipping the solution whenever it falls outside the admissible control range. The algorithm terminates once the objective improvement satisfies $\left| \mathcal{J}(\mathbf{u}^{\mathrm{opt}}_{t+1})
      - \mathcal{J}(\mathbf{u}^{\mathrm{opt}}_t) \right| < \delta$ for a predefined tolerance
$\delta$. The default values of all hyperparameters in Algorithm~\ref{alg:kernelSOS-MPPI} are summarized in Table~\ref{tab:hyperparameters}.

\begin{algorithm}[t]
\caption{Global-MPPI algorithm}
\label{alg:kernelSOS-MPPI}
\begin{algorithmic}[1]

\Statex \textbf{Input:} $\mathbf{u}_{0}, \sigma_{lse}, \Delta, \rho, \gamma$
\Statex \textbf{Notation:} $\mathbf{u}_t:= (u_t,\dots,u_{t+H-1})$ denotes a control sequence over a horizon of length $H$; 
\State $\mathbf{u}^{\mathrm{opt}}_{t} \leftarrow \mathbf{u}_{0}$
\For{$t = 0$ \textbf{to} T}
    \For{$j = 0$ \textbf{to} $N_{\mathrm{restart}}$}
        % \State  $\mathbf{u}^{(i)}_t$ \sim $\mathcal{U}\!\left(\mathbf{u}^{opt}_t -       \Delta,\; \mathbf{u}^{opt}_t + \Delta\right)$ $\quad i \in [N]$
        \State $\mathbf{u}^{(i)}_t \sim \mathrm{Unif}\!\left(
        \mathbf{u}^{\mathrm{opt}}_t - \Delta,\;
        \mathbf{u}^{\mathrm{opt}}_t + \Delta
        \right), \quad i \in [N]$
        \State {\color{green!50!black}\ttfamily // graduated non-convexity:}
        % \State Rollout $u^{(i)}_{t:t+H}$ to evaluate $J(u^{(i)}_{t:t+H})$
        \State $\mathcal{J}_{{lse}}(\mathbf{u}^{(i)}_t) \leftarrow \operatorname{LSEsmooth}(\mathbf{u}^{(i)}_t, \sigma_{lse}), i \in [N]$ 
        \State Define dataset $\mathcal{D} = \{\mathbf{u}^{(i)}_t, \mathcal{J}_{{lse}}(\mathbf{u}^{(i)}_t)\}^N_{i=1}$
        % \Statex \hspace{\algorithmicindent} \hspace{\algorithmicindent}
        % $f_{\sigma_{lse}}(\{u^{(i)}_{t:t+H}\}_{i=1}^{N}, J(\{u^{(i)}_{t:t+H}\}_{i=1}^{N}))$
        \State {\color{green!50!black}\ttfamily // global search:}
        \State $\sigma_{ksos} \gets \operatorname{AutoCalibration}(\mathcal{D})$
        \State $\mathbf{u}^{\mathrm{ref}}_t \gets \operatorname{KernelSOS}(\mathcal{D}, \sigma_{ksos})$
        \State {\color{green!50!black}\ttfamily // local refinement:}
        \State $\mathbf{u}^{\mathrm{opt}}_t \gets \operatorname{MPPI}(\mathbf{u}^{\mathrm{ref}}_t, \sigma_{lse})$
        \State {\color{green!50!black}\ttfamily // update hyperparameters:}
        \State $\sigma_{lse} \leftarrow \rho\,\sigma_{lse},\;\Delta \leftarrow \gamma\,\Delta$
    \EndFor
    % \mathbf{u}^{\mathrm{opt}}_t \leftarrow \mathrm{MPPI}_{\sigma_{lse}}(\mathbf{u}^{\mathrm{ref}}_t)
    \State {\color{green!50!black}\ttfamily // receding horizon:}
    \State $\mathbf{u}^{\mathrm{opt}}_{t+1} \leftarrow \text{shift}(\mathbf{u}^{\mathrm{opt}}_t)$ 
    \If{$\left| \mathcal{J}(\mathbf{u}^{\mathrm{opt}}_{t+1})
      - \mathcal{J}(\mathbf{u}^{\mathrm{opt}}_{t}) \right| < \delta$}
    \State \textbf{break}
    \EndIf
\EndFor

\end{algorithmic}
\end{algorithm}

\begin{table}[t]
    \centering
    \caption{Definition of Hyperparameters in Global-MPPI}
    \label{tab:hyperparameters}
    \begin{tabular}{c l}
        \hline
        \textbf{Parameters} & \textbf{Descriptions} \\
        \hline
        $\mathrm{Unif}$    & Uniformly sampling \\
        $N$      & Number of KernelSOS samples: 80 \\
        $N_{lse}$    & Number of LSE samples: 100 \\
        $N_{\mathrm{MPPI}}$      & Number of MPPI samples: 256 \\
        $N_{\mathrm{restart}}$    & Number of restart: 5 \\
        $\sigma_{lse}$    & Initial LSE noise: $0.4$ \\
        $\mathbf{u}_0$    & Initial control sequence: $0.5\left(\mathbf{u}_{\max}+\mathbf{u}_{\min}\right)$ \\
        $\mu$    & KernelSOS regularization: $1e^{-5}$ \\
        $k_\sigma$    & KernelSOS kernel: Laplace \\
        $\Delta$    & Initial sampling radius: $0.5\left(\mathbf{u}_{\max}-\mathbf{u}_{\min}\right)$ \\
        $\rho$    & \ac{LSE} shrink factor: $0.5$ \\
        $\gamma$    & Sampling radius shrink factor: $0.85$ \\
        $\delta$ & Converge tolerance: $1e^{-4}$ \\
        \hline
    \end{tabular}
\end{table}
\section{Experiments}
\label{experiments}
In this section, we evaluate Global-MPPI on two contact-rich benchmark tasks~\cite{kurtz2024hydrax}: the PushT task and dexterous in-hand manipulation. We compare the performance of our approach against several state-of-the-art sampling-based optimization methods, including Predictive Sampling \cite{jordana2025introduction}, vanilla MPPI~\cite{alvarez2025real}, and DIAL-MPC~\cite{xue2025full}, to demonstrate its ability to explore high-quality solutions in non-smooth, non-convex optimization landscapes. % All code will be open-sourced upon acceptance of the manuscript to support reproducibility.
\subsection{Experiment setting}
Our experiments were conducted in MuJoCo, a GPU-accelerated physics simulator. We ran the benchmarks on a workstation equipped with an Intel i9 CPU and an RTX 4090 GPU. To ensure fair comparisons, all methods were tested in the same environments, initialized from the same states, and run with the same seeds. Unless otherwise specified, Global-MPPI uses the same hyperparameters for all experiments as summarized in Table~\ref{tab:hyperparameters}. In our implementation, sampling, rollouts, function evaluation, \ac{LSE} smoothing, and \ac{MPPI} updates are executed on the GPU to leverage parallel computation, while KernelSOS is performed on the CPU. We use a custom interior-point solver for SDP to exploit the rank-one structure of the KernelSOS constraints~\cite{rudi2020finding}, which is computationally faster than off-the-shelf SDP solvers such as MOSEK~\cite{aps2019mosek} or SCS~\cite{flicker2020comparison}. 
\subsection{PushT task}
The PushT task is a planar contact-rich manipulation task in which a robotic pusher moves a T-shaped object to a specified target pose. The task involves discontinuous contact dynamics, frictional interactions, and frequent contact mode switches. Consequently, the optimization landscape is highly non-smooth and non-convex. As a result, small perturbations in the control sequence can lead to qualitatively different object motions, turning the PushT task into a particularly challenging planning problem.

We evaluate the performance of Global-MPPI on the PushT task. The system state $x_t$ consists of both the object pose and the pusher pose, and the control input $u_t$ corresponds to the pusher’s position. Starting from an initial configuration, the objective is to compute a control sequence for the robot pusher that drives the object to a desired target pose. We define the cost function as a distance metric on the $\mathbb{R}^2 \times \mathrm{SO}(2)$ manifold, penalizing both translational and rotational errors:
\begin{equation}
\begin{aligned}
\mathcal{J}
&=
\sum_{t=0}^{T-1}
\underbrace{
\Delta t
\Big(
\| p_t - p_g \|^2
+ w
\big\| \log( q_g^{-1} \otimes q_t ) \big\|^2
\Big)
}_{\ell_t(\cdot)}
\\
&\quad
+
\underbrace{
\| p_T - p_g \|^2
+ w
\big\| \log( q_g^{-1} \otimes q_T ) \big\|^2
}_{\ell_T(\cdot)}.
\end{aligned}
\end{equation}
Here, the object pose at time $t$ is represented by $(p_t, q_t) \in \mathbb{R}^2 \times \mathrm{SO}(2)$, and the target pose is $(p_g, q_g) \in \mathbb{R}^2 \times \mathrm{SO}(2)$. $\Delta t$ denotes the time step, and the weight $w$ is used to balance translational and rotational errors. The operator $\otimes$ denotes rotation composition, and $\log(\cdot)$ maps the relative rotation $q_g^{-1} \otimes q_t$ to the Lie algebra of $\mathrm{SO}(2)$, from which we extract a scalar rotation error. 

\begin{figure}[!t]
    \centering
    \includegraphics[width=1\linewidth]{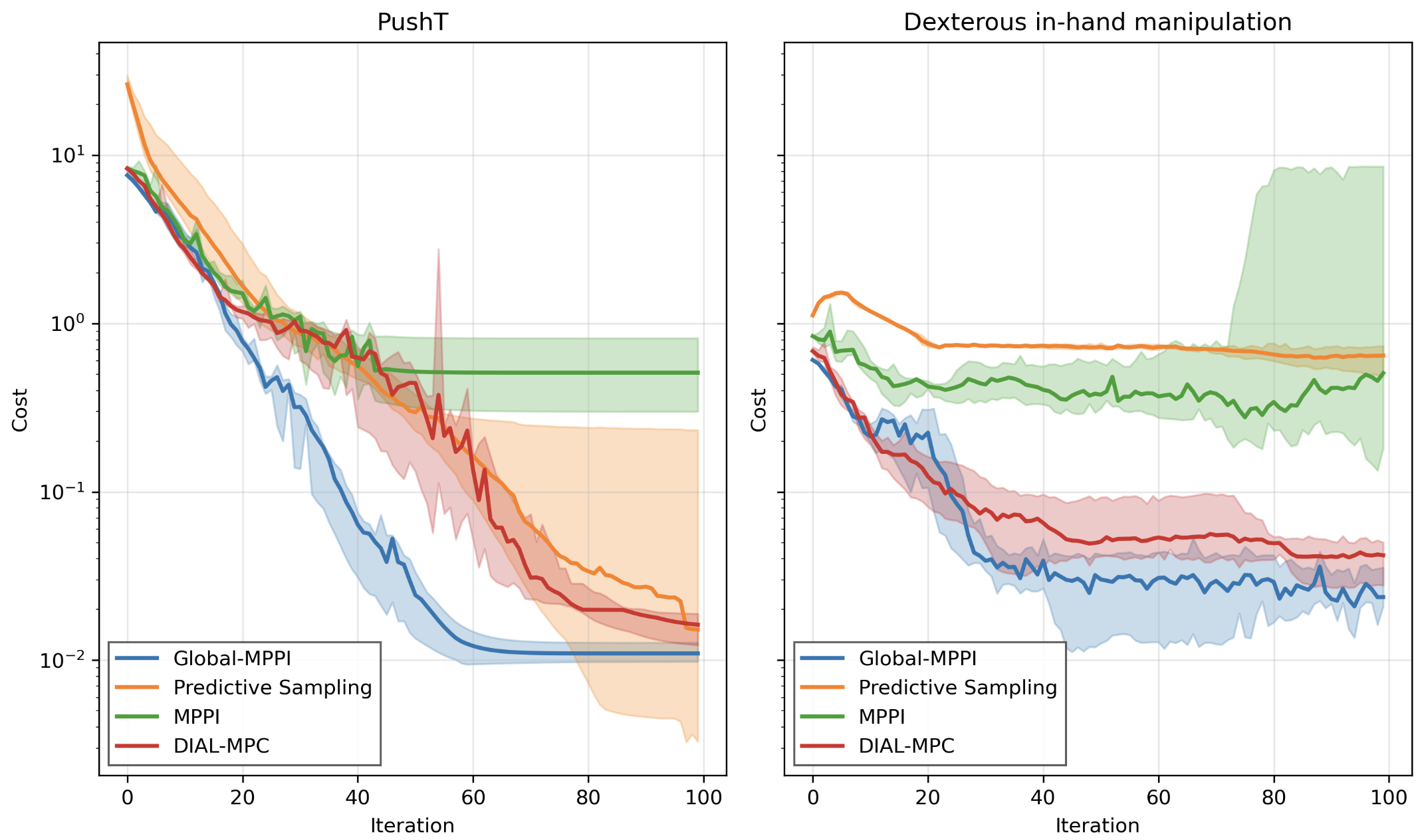}
    \caption{Cost convergence comparison for PushT and dexterous in-hand manipulation task. The solid lines represent the median across six random seeds, and the shaded regions indicate the interquartile range. Our method achieves faster convergence, lower final costs, and reduced variance compared to other sampling-based methods.}
    \label{fig:pusht_cost_convergence}
\end{figure}

\begin{figure}[!t]
    \centering
    \includegraphics[width=1\linewidth]{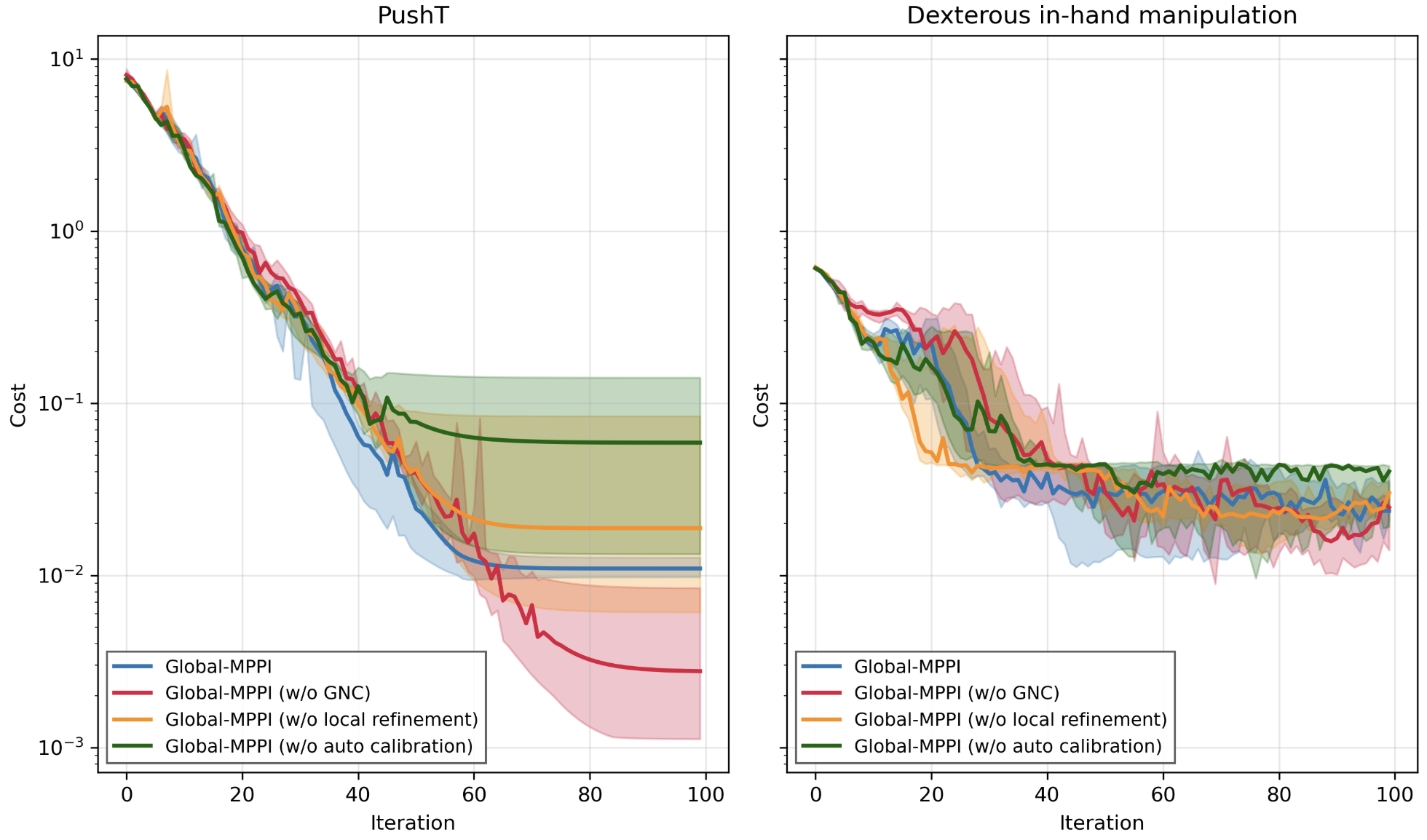}
    \caption{Ablation study of Global-MPPI on the PushT and dexterous in-hand manipulation tasks. We evaluate Global-MPPI by removing key components, including GNC, local refinement, and automatic calibration. The solid lines represent the median across six random seeds, and the shaded regions indicate the interquartile range. For the PushT task, removing any of these components significantly degrades performance, leading to slower convergence, higher final costs, and increased variance. The dexterous manipulation task is less sensitive to removing the refinement or GNC, but auto calibration remains important.} %In contrast, the full Global-MPPI consistently achieves faster convergence, lower final costs, and more stable performance across random seeds.}
    \label{fig:global-mppi_compare}
\end{figure}

In our PushT experiments, we set the weight $w = 0.3$ and the time step $\Delta t = 0.01$ over a 1-second planning horizon. We use six control points for the cubic spline parameterization, resulting in a 12-dimensional control sequence $\mathbf{u}$. The benchmark results are summarized in the left panel of Fig.~\ref{fig:pusht_cost_convergence}. Global-MPPI outperforms baseline methods, achieving faster convergence, lower final costs, and reduced variance. MPPI and DIAL-MPC, as more local optimization methods, often become trapped in poor local minima and sometimes struggle to reduce the cost. Predictive sampling is more robust to local minima but converges more slowly and with higher variance. We also conduct an ablation study of Global-MPPI by removing GNC, local refinement, and automatic calibration. The results are summarized in the left panel of Fig.~\ref{fig:global-mppi_compare}. Removing local refinement or automatic calibration leads to higher final costs and increased variance. Although removing GNC yields a slightly lower final cost, incorporating GNC improves convergence speed and stability, suggesting that adaptive GNC strategies warrant further investigation.

% \begin{figure}[!th]
%     \centering
%     \includegraphics[width=1\linewidth]{figs/pusht_cost_convergence_comparison.png}
%     \caption{Cost convergence comparison for PushT. The solid lines represent the median across six random seeds, and the shaded regions indicate the interquartile range. Global-MPPI converges faster and achieves lower final costs compared with other sampling-based methods.}
%     \label{fig:pusht_cost_convergence}
% \end{figure}

\begin{figure}[!t]
    \centering
    \includegraphics[width=1\linewidth]{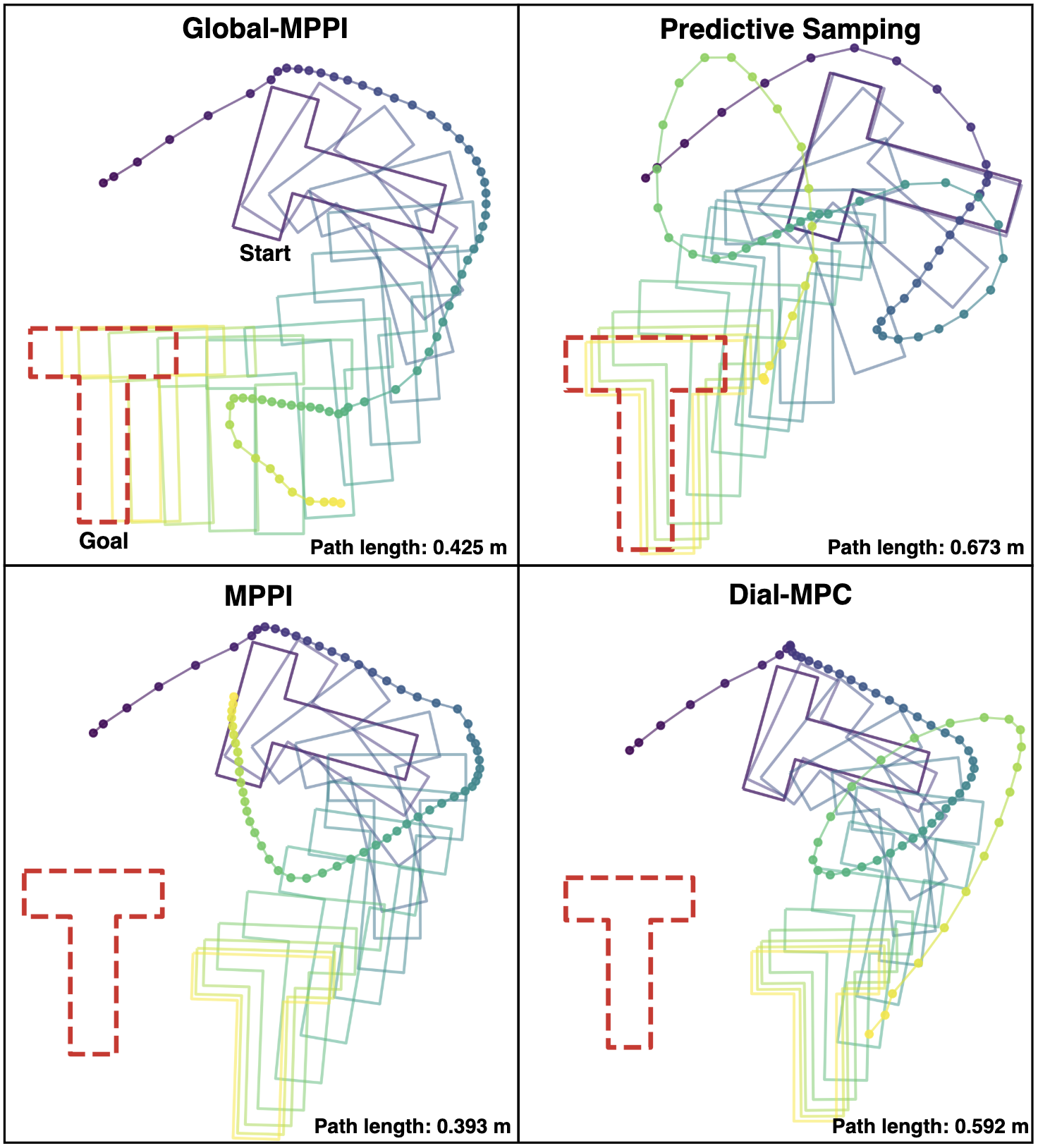}
    \caption{Visualization of the PushT task with four different methods. The dashed red region denotes the target pose of the object. Under the same environment, Global-MPPI achieves the lowest cost and shortest trajectory path length. Predictive sampling can push the object into the target region, but it incurs large stage costs and inefficient trajectories due to random exploration. MPPI and DIAL-MPC, as local optimization methods, fail to push the object to the target region.}
    \label{fig:pusht_task_path}
\end{figure}

The robot trajectories produced by different methods in the PushT task are shown in Fig.~\ref{fig:pusht_task_path}, where the dashed red region denotes the target pose of the T-shaped object. Under the same initial and goal configurations, Global-MPPI achieves shorter trajectories, faster convergence, and lower final costs, outperforming the baseline methods. Notably, although the cost function does not explicitly penalize path length, Global-MPPI consistently identifies low-cost actions that efficiently reduce the task objective, yielding shorter, more efficient robot trajectories. In contrast, predictive sampling can successfully push the object into the target region, but incurs higher intermediate costs and generates less efficient trajectories. MPPI and DIAL-MPC, as local optimization methods, become trapped in suboptimal local minima and fail to move the object to the target region.

\subsection{Dexterous in-hand manipulation}

We further evaluate our approach on a high-dimensional dexterous in-hand manipulation task, which requires a multi-fingered hand to rotate and stabilize a cube to a desired target orientation while maintaining a stable grasp. This task involves intermittent multi-point contacts, frequent contact-mode switches, and highly nonlinear dynamics, resulting in a complex optimization landscape with many local minima. The cost function for this task is defined as a distance metric on the $\mathrm{SO}(3)$ manifold:

\begin{equation}
\begin{aligned}
\mathcal{J}
&=
\sum_{t=0}^{T-1}
\underbrace{
\Delta t
\Big(
\big\| \log( q_g^{-1} \otimes q_t ) \big\|^2
\Big)
}_{\ell_t(\cdot)}
+
\underbrace{
 \
\big\| \log( q_g^{-1} \otimes q_T ) \big\|^2
}_{\ell_T(\cdot)}.
\end{aligned}
\end{equation}
where the object orientation at time $t$ is denoted by $q_t \in \mathrm{SO}(3)$, and the target orientation is $q_g \in \mathrm{SO}(3)$.

Due to the small-scale joint motions of the dexterous hand, we parameterize the control trajectory using four control points over a 0.25-second horizon. We use a 16-DoF LEAP hand~\cite{shaw2023leaphand} for this dexterous in-hand manipulation experiment. The dimension of the control sequence $\mathbf{u}$ is 64. The right panel of Fig.~\ref{fig:pusht_cost_convergence} shows the cost convergence comparison across different methods. Global-MPPI achieves the fastest convergence and the lowest final cost, demonstrating robust performance in the high-dimensional contact-rich problem. Predictive sampling shows limited improvement and plateaus at a significantly higher cost due to inefficient exploration in the high-dimensional action space. MPPI frequently converges to suboptimal local minima or fails to maintain stable object manipulation, resulting in inferior final performance. DIAL-MPC demonstrates improved performance compared with the PushT task; however, it attains higher final costs than Global-MPPI and converges slower, with Global-MPPI reaching the final best performance of DIAL-MPC already after less than 40 iterations, effectively halving the iteration time. The ablation results are shown in the right panel of Fig.~\ref{fig:global-mppi_compare}. Although all methods achieve almost similar final costs, the full Global-MPPI converges faster, reaching the final cost within approximately 30 iterations.

% \textcolor{red}{more analysis for global-MPPI w/o local refinement and calirbation}
% Global-MPPI without LSE converges more slowly and exhibits higher variance, highlighting the importance of LSE-based graduated non-convexity for stabilizing global surrogate construction. 

% \begin{figure}[!th]
%     \centering
%     \includegraphics[width=1\linewidth]{figs/Dexterous hand.png}
%     \caption{Cost convergence comparison for dexterous in-hand manipulation.}
%     \label{fig:dexterous_cost_convergence}
% \end{figure}

% \addtolength{\textheight}{-4cm}   % This command serves to balance the column lengths
                                  % on the last page of the document manually. It shortens
                                  % the textheight of the last page by a suitable amount.
                                  % This command does not take effect until the next page
                                  % so it should come on the page before the last. Make
                                  % sure that you do not shorten the textheight too much.
% \balance
\section{Conclusion}
\label{Conclusion}
We presented Global-MPPI, a global sampling-based trajectory optimization framework tailored for contact-implicit trajectory optimization in contact-rich manipulation problems. By combining graduated non-convexity, kernelized global surrogates, and efficient local refinement, Global-MPPI enables global exploration and robust local convergence toward high-quality solutions.

Despite its effectiveness, the proposed approach has several limitations. First, scalability remains a challenge. As the problem dimension and planning horizon increase, the computational cost of \ac{SDP} scales cubically with the number of samples. Future improvements can explore GPU-accelerated SDP solvers and efficient sampling to further advance real-time and large-scale applications.

Second, our approach relies on a parallelizable physics-based simulator to rollout trajectories and evaluate sample costs. Therefore, the performance of our approach depends on both the fidelity and computational efficiency of the simulator. Inaccurate modeling of complex dynamics may introduce a sim-to-real gap when deploying our approach on real-world systems. In practice, this issue can be partially mitigated by the receding-horizon MPC framework, which enables frequent re-planning and helps reduce the impact of model mismatch. We leave a comprehensive real-world implementation and validation to future work.
% \IEEEtriggeratref{34}
\bibliographystyle{IEEEtran}
% \enlargethispage{10\baselineskip}
% \IEEEtriggercmd{\enlargethispage{-5}}
% \balance
% \vspace{-20mm}
\bibliography{ref,ref-fd}
%%%%%%%%%%%%%%%%%%%%%%%%%%%%%%%%%%%%%%%%%%%%%%%%%%%%%%%%%%%%%%%%%%%%%%%%%%%%%%%%
% \section*{APPENDIX}

% Appendixes should appear before the acknowledgment.

% \section*{ACKNOWLEDGMENT}

% The preferred spelling of the word ÒacknowledgmentÓ in America is without an ÒeÓ after the ÒgÓ. Avoid the stilted expression, ÒOne of us (R. B. G.) thanks . . .Ó  Instead, try ÒR. B. G. thanksÓ. Put sponsor acknowledgments in the unnumbered footnote on the first page.
%%%%%%%%%%%%%%%%%%%%%%%%%%%%%%%%%%%%%%%%%%%%%%%%%%%%%%%%%%%%%%%%%%%%%%%%%%%%%%%%

\end{document}